\documentclass[letterpaper, 10 pt, journal, twoside]{IEEEtran}

\usepackage{graphics} 
\usepackage{epsfig} 
\usepackage{mathptmx} 
\usepackage{times} 
\usepackage{amsmath} 
\usepackage{amssymb}  

\usepackage{siunitx}
\usepackage{hyperref}
\usepackage{comment}
\usepackage{float}
\usepackage{array}
\usepackage{multirow}
\usepackage{booktabs}
\usepackage[table]{xcolor}

\usepackage{algorithm}
\usepackage{algpseudocode}
\algnewcommand\algorithmicforeach{\textbf{for each}}
\algdef{S}[FOR]{ForEach}[1]{\algorithmicforeach\ #1\ \algorithmicdo}
\DeclareMathOperator*{\argmax}{argmax} 
\DeclareMathOperator{\arctantwo}{arctan2}

\newcommand{\etal}{et~al.\ }
\newcommand{\eg}{e.\,g.,\ }
\newcommand{\ie}{i.\,e.,\ }

\begin{document}

\title{
Keypoints-Based Deep Feature Fusion for Cooperative Vehicle Detection of Autonomous Driving
}
\author{Yunshuang Yuan$^{1}$, Hao Cheng$^{1}$ and Monika Sester$^{1}$

\thanks{$^{1}$Yunshuang Yuan, Hao Cheng, and Monika Sester are with the Institute of Cartography and Geo-informatics, Leibniz University Hannover, Germany
{\tt\footnotesize \{firstname.lastname\}@ikg.uni-hannover.de}}%

}


\maketitle


\begin{abstract}

Sharing collective perception messages (CPM) between vehicles is investigated to decrease occlusions so as to improve the perception accuracy and safety of autonomous driving.
However, highly accurate data sharing and low communication overhead is a big challenge for collective perception, especially when real-time communication is required among connected and automated vehicles.
In this paper, we propose an efficient and effective keypoints-based deep feature fusion framework built on the 3D object detector PV-RCNN, called \textit{Fusion} PV-RCNN (\textbf{FPV-RCNN} for short), for collective perception.
We introduce a high-performance bounding box proposal matching module and a keypoints selection strategy to compress the CPM size and solve the multi-vehicle data fusion problem. 
Besides, we also propose an effective localization error correction module based on the maximum consensus principle to increase the robustness of the data fusion.
Compared to a bird's-eye view (BEV) keypoints feature fusion, \textbf{FPV-RCNN} achieves improved detection accuracy by about 9\% at a high evaluation criterion (IoU 0.7) on the synthetic dataset COMAP dedicated to collective perception.  
In addition, its performance is comparable to two raw data fusion baselines that have no data loss in sharing.
Moreover, our method also significantly decreases the CPM size to less than 0.3KB, and is thus about 50 times smaller than the BEV feature map sharing used in previous works. 
Even with further decreased CPM feature channels, \ie~from 128 to 32, the detection performance does not show apparent drops.
The code of our method is available at \url{https://github.com/YuanYunshuang/FPV\_RCNN}.

\end{abstract}

\section{INTRODUCTION}\label{intro}
\IEEEPARstart{U}{nderstanding} the surrounding environment is one of the most important tasks of autonomous driving, especially for those automated vehicles (AV) driving in complex real-world situations. 
Such an AV is normally equipped with different sensors like cameras, LiDARs, and Sonars in order to sense the world \cite{robo_perception}. 
However, perceiving the environment only using the data collected by these sensors mounted on a single ego vehicle has many limitations, such as occlusion, limited sensor observation range, and noise. 
In this regard, cooperative perception based on connected and automated vehicles (CAVs) can effectively mitigate these problems by sharing sensed information collected from different viewing directions of multiple AVs in a network.
The perceived information is shared among vehicles via Collective Perception Messages (CPMs). 
In this way, the accuracy and reliability requirements of the sensors on each vehicle can be relaxed, and therefore the price of each AV is lowered as well~\cite{Shan2021DemoCP}. 
However, the challenging part of cooperative perception is defining the information to be shared and fusing the shared information via a limited communication network bandwidth. 
Hence, the goal is to obtain the best perception performance with the least data transmission in the network of cooperative agents.

\begin{figure}[t]
\begin{center}
    \includegraphics[trim=0in 0in 0in 0.1in, clip=true, width=0.5\textwidth, height=6.5cm]{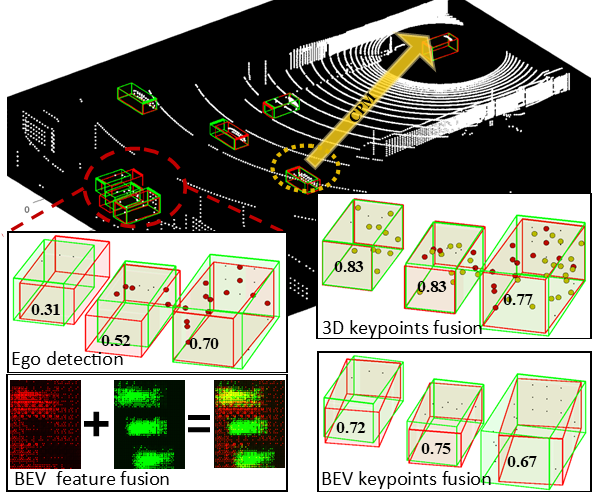}
    \caption{{The detection result of an exemplary frame with two CAVs}. The vehicle in the yellow dashed circle shares CPM to the ego vehicle (upper right). According to the IoUs (marked in the boxes) against the ground truth, our proposed method of the 3D keypoints fusion outperforms the BEV keypoints fusion by a large margin for improving the ego vehicle's detection.}
\label{fig:result_show}
\end{center}
\end{figure}

Accurate data sharing and low communication overhead is still a bottleneck for cooperative vehicle detection demanding real-time communication in autonomous driving.
In theory, sharing raw data gives the best performance because no information is lost.
But this can easily congest the communication network with heavy data loads. 
In contrast, sharing the fully processed data, \eg~detected objects, needs fewer communication resources. 
Nevertheless, object-wise fusion is very sensitive to the localization noise of the agents. 
Matching the detected objects coming from different agents can be very difficult, especially those that are inaccurately detected by distant sensors. 
As a trade-off, deep features extracted by deep neural networks from the raw data can decrease the amount of data to be shared and at the same time maintain a relatively high performance of data fusion. 
Previous works ~\cite{fcooper,marvasti2020feature,v2vnet} achieve this by contracting bird's-eye view (BEV) deep features maps which are, however, very sparse and can be further compressed to avoid redundancy. Moreover, due to the low resolution, fusing such feature maps may even fail to predict accurate bounding boxes.
To this end, this paper proposes a more robust deep feature sharing and fusion framework by extending the established framework PV-RCNN~\cite{pvrcnn} to colletive perception scenarios. 
Our framework uses PointNet~\cite{Qi2017PointNetDL} and point set abstraction~\cite{pointnet_pp} to aggregate the information from multi-scale receptive fields for the selected high accurate 3D keypoints from different point clouds, which are then shared and fused to generate more accurate detection.
In comparison to the BEV keypoints fusion, with reduced communication overhead our 3D keypoints fusion still achieves higher detection accuracy. An example tested on the synthetic collective perception dataset COMAP \cite{comap} is shown in Fig.~\ref{fig:result_show}.

Our main contributions are summarized as follows:
\begin{itemize}
    \item[1)] We propose a 3D keypoints feature fusion scheme for cooperative vehicle detection to remedy the problem of low bounding box localization accuracy of the schemes that are based on the BEV feature fusion. 
    \item[2)] We introduce a keypoints selection module to reduce the redundancy of shared deep features so as to decrease the communication overhead.  
    \item[3)] We propose an efficient and robust localization correction module and a bounding box matching module that can generate bounding box proposals of high quality for the deep feature fusion in the later stage.
    \item[4)] Our proposed method not only outperforms the state-of-the-art method that uses BEV feature fusion for collective perception with a large margin but also reduces the CPM data size by a large scale. 
\end{itemize}

\begin{figure*}[t]
\begin{center}
    \includegraphics[trim=0.01in 0in 0in 0in, clip=true,width=0.8\textwidth,height=5.7cm]{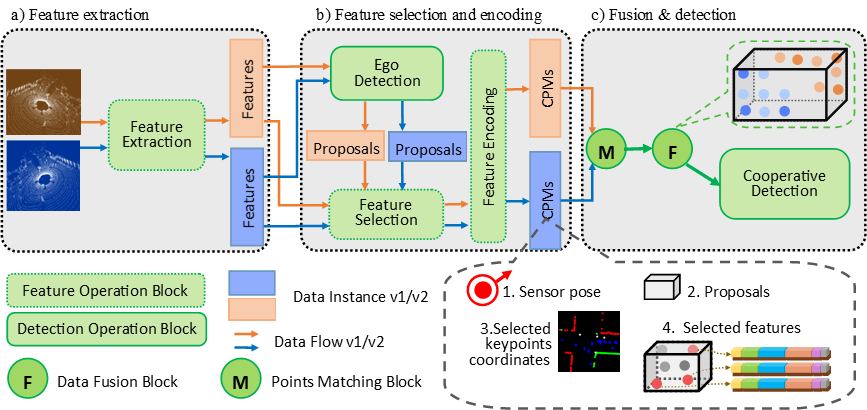}
    \caption{An overview of the keypoints deep feature fusion framework (\textbf{FPV-RCNN}). 
    }
\label{fig:fusion_pipeline}
\end{center}
\end{figure*}
\section{Related Work}\label{related_works}

In general, cooperative perception can be achieved by means of Vehicle-to-Infrastructure (V2I) and Vehicle-to-Vehicle (V2V) communication.
V2I communication offers the opportunity to exchange sensory information between an ego vehicle and the infrastructure. This helps the ego vehicle go beyond the limitations of its own perception system. 
A successful application by Yang~\etal~\cite{cosmos} is the so-called smart intersection, where information for object detection and tracking is shared via the BEV observation from the static cameras at the intersection to the ego vehicle. 
These cameras are easy to deploy, whereas their perceptions are limited to traffic scenarios at the specific intersection. 
In contrast, V2V communication is not limited to the defined location. 
In a CAV network, each vehicle can be seen as a node with multiple sensors, the sensed data can be shared across the vehicles at any place~\cite{v2i_1}. 
Our work focuses on V2V communication for object detection.

In V2V communication, different approaches are proposed to communicate the data in a CAV network. 
In this paper, we sort data fusion strategies as (1) raw data sharing, (2) fully processed data, such as detected objects, and (3) half-processed data. 
The study by Marvasti~\etal\cite{marvasti2020feature} shows that raw data sharing provides rich information for object detection. 
It, however, consumes large bandwidth and is not feasible for autonomous driving that requires real-time communication. Contrary to raw data sharing, \cite{fdf2019,fdf2020,cooper}~propose to only share the detected objects for more efficient communication. 
However, the work by Wang~\etal\cite{v2vnet} has shown that this late fusion of fully processed data performs worse than either early raw data fusion or half-processed data fusion. 

In order to reduce communication resource consumption without a compromise of performance, sharing half-processed data is further explored. 
In an extreme case, the objects mis-detected by all independent sensors can be detected after this data fusion~\cite{fcooper}. 
For example, instead of fusing the detected objects, Chen~\etal\cite{fcooper} extend their previous work~\cite{cooper} by fusing voxel features and deep features learned using a Deep Neural Network (DNN) for cooperative perception. 
On the one hand, significant performance improvement has been shown on the real-world datasets KITTI~\cite{kitti} and T\&J~\cite{cooper} only for dedicated traffic scenarios, \eg in a parking lot~\cite{cooper}. 
On the other hand, these datasets are not dedicated to collective perception but rather to a single egocentric perspective. 
This is because collective perception requires multiple vehicles to share a certain degree of field-of-view (FOV) at the same time. 
But acquiring such a real-world dataset not only needs expensive equipment but also needs numerous hours of manual labeling to obtain ground truth information. 
Therefore, many recent works~\cite{marvasti2020feature,lidarsim} resort to synthetic data for a more comprehensive empirical study. 
Data generator and simulation tools \eg~CARLA and SUMO~\cite{sumo}, can not only be manipulated to generate a large amount of realistic data in various traffic situations for cooperative perception, but also provide accurate ground truth information. 
In~\cite{marvasti2020feature}, the comparison of different data fusion strategies on a simulated point cloud dataset generated by CARLA indicates that both the raw data and deep feature fusion outperform the object-wise fusion by a big margin, especially when vehicle localization errors are introduced. 
In addition, \cite{v2vnet} also confirms that on the simulated dataset LiDARsim~\cite{lidarsim}, sharing compressed deep feature maps achieves high accurate object detection while satisfying communication bandwidth requirements. 

Despite the preliminary success of deep feature fusion, the shared feature maps still contain too much redundancy due to their sparsity. 
These deep features are highly abstract, which are difficult to be selected, compressed, and finally fused by a neural network. 
For example, \cite{fcooper,marvasti2020feature,v2vnet} tried to share intermediate feature maps for vehicle detection. 
It was found that this strategy is not robust in providing highly accurate bounding box predictions because the shared feature maps are of low resolution, \ie~8$\times$ down-sampled from the raw data. 
Besides, all previous deep feature fusion frameworks mentioned above evaluate their performance with object-wise fusion without localization error correction and have not provided the implementation details of the object-wise fusion method. 
However, different implementations of local object detection and object-wise fusion can greatly influence the final result. 
Moreover, the localization error can be recovered without much effort only by the geometry of the detected vehicles in most of the situations as far as two matchings of the detected vehicles between the ego and cooperative vehicles are available. 
Hence, it is also important to analyze the final fusion result with localization error correction, which is proposed in this paper.

To summarize, instead of sharing deep features, we investigate sharing only the selected keypoint features, aiming to further reduce the feature size while keeping the performance for object detection. 
Moreover, we also introduce localization errors and error correction to guarantee a fair comparison of the performance of all fusion methods.

\section{Method}
\label{sec:method}
\subsection{Problem formulation}\label{sec:formulation}
We formulate the collective perception problem in an egocentric way. 
Within a communication range $R_\text{c}$ of the ego-vehicle $C_0$, $N_\text{v}$ number of cooperative CAVs $\{C_1, C_2, \ldots C_{N_\text{v}}\}$ as well as the ego CAV have generated the point cloud set $\textbf{PC}=\{PC_0, PC_1, \ldots, PC_{N_\text{v}}\}$ at time $t$. 
The bounding boxes (BBoxes) of the $N_i$ vehicles detected based on $PC_i$ are called proposals and notated as $B_i=\{(b_j, s_j)\mid j=1,\ldots,N_i)\}$. 
Each instance in $B_i$ is a pair which contains one detected vehicle $b_j = (x, y, z, w, l, h, r)$ and its corresponding detection confidence $s_j$. 
In this notation, $xyz$ indicates the BBox center, $wlh$ the dimensions, and $r\in[-\pi, \pi]$ the orientation. In our proposed framework, the cooperative CAV $C_i$ ($1\leq i\leq N_\text{v}$) generates and shares to the ego CAV $C_0$ the $CPM_i$ that contains $B_i$, the selected and aggregated deep feature information $F_i$ and the coordinates of $K_i$ keypoints for localization error correction. 
Then ego vehicle $C_0$ fuses the information of the received CPMs with the local information and generates the final refined predictions of the BBoxes.

\subsection{3D Keypoints deep features fusion}
The fusion framework proposed is built based on the 3D object detector PV-RCNN~\cite{pvrcnn}, hence we term it as \textit{Fusion} PV-RCNN, or \textbf{FPV-RCNN} for short in the rest of this paper.
Figure~\ref{fig:fusion_pipeline} demonstrates a fusion example of two CAVs. 
It is also straightforward to extend this framework to an arbitrary number of CAVs. 
As depicted in the figure, data flows of the two CAVs are colored blue and yellow, respectively.
We first extract deep features separately from point clouds (Fig. \ref{fig:fusion_pipeline}a) and then select and encode the most important features for sharing (Fig. \ref{fig:fusion_pipeline}b). 
At last, the shared features are fused for the final detection (Fig. \ref{fig:fusion_pipeline}c).

\begin{figure}[t]
\begin{center}
    \includegraphics[trim=0.01in 0.01in 0in 0.2in, clip=true,width=0.45\textwidth]{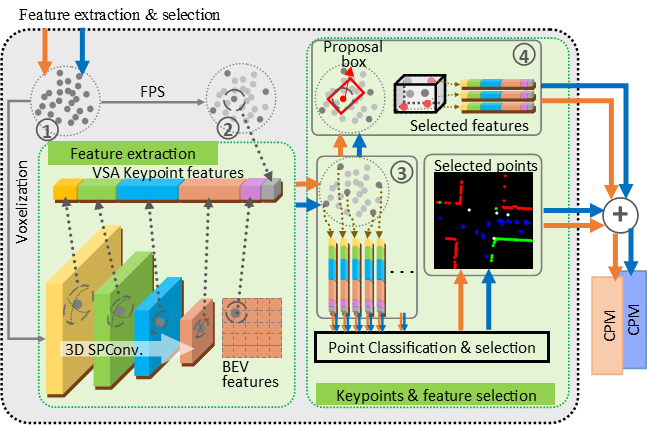}
    \caption{Feature extraction and selection. 
    }
\label{fig:feature_selection}
\end{center}
\end{figure}

\paragraph{Feature extraction}
To extract the 3D features of point clouds, we adopt a voxel-based sparse CNN backbone network from \cite{pvrcnn} because of its high efficiency and accuracy. 
This network is demonstrated in the bottom left of Fig.~\ref{fig:feature_selection}. 
The raw point cloud is first voxelized and then passed to a block of 3D sparse convolutions \cite{Graham2015Sparse3C, Graham2017SubmanifoldSC}. 
The original voxel features are encoded and 8$\times$ down-sampled to 3D deep features. 
The features from the last sparse convolution layer are then compressed and projected to BEV features.

\paragraph{Feature selection and encoding}
The ego-detection module adopts the detection head from CIA-SSD~\cite{CIASSD} since it has a simple structure and can generate better proposals than the proposal generation module in PV-RCNN. 
Besides, CIA-SSD calibrates the detection scores with IoUs which is critical for our matching in Algorithm~\ref{alg:cbp} that uses the scores for merging. 
This module generates proposals $B_i$ which are then utilized for selecting feature points.
Only the feature points inside the proposals are selected, further encoded, and compressed to the CPM format to decrease the CPM size.

The details of the feature selection are shown in Fig.~\ref{fig:feature_selection}.
Furthest Points Sampling (FPS) is used to sample a pre-defined number $N_\text{kpts}$ of evenly distributed sparse keypoints (step \textcircled{1} to \textcircled{2}). 
Based on the selected keypoints in \textcircled{2}, the Voxel Set Abstraction (VSA) module with the same parameters is adopted from \cite{pvrcnn} to aggregate deep features for each selected keypoint. 
This module aggregates neighboring voxel-wise features of different resolutions and abstract levels for each keypoints with a PointNet~\cite{Qi2017PointNetDL}. 
The aggregated keypoint features are then split into two paths. 
On the first path, these points are further down-sampled by only selecting the keypoints that are inside the proposal $B_i$ (step \textcircled{3} to \textcircled{4}) for generating CPMs. 
On the second path, they are classified and selected for localization error correction. 
For the point cloud $PC_i$, we compose the $CPM_i$ with the sensor pose of CAV $C_i$, proposals $B_i$, coordinates and features of keypoints $F_i$ for fusion and $K_i$ keypoints coordinates for localization error correction, as shown in the dash-line box in Fig.~\ref{fig:fusion_pipeline}.

\paragraph{Fusion and detection}
In the fusion step, the ego-vehicle transforms all received proposal boxes and keypoints to the same local coordinate system. 
The transformed proposals are then clustered and merged using algorithm \ref{alg:cbp}. 
If the IoU of two proposals in set $\mathcal{B}$ is above a pre-defined threshold (\eg 0.3), they are clustered into the same subsets $C_k$ (step~\ref{alg:clustering_s}-\ref{alg:clustering_e}). 
In each $C_k$, we first align the direction $r_i$ of each BBox $b_i$ to the dominant direction of all BBoxes in this cluster in order to prevent erroneous orientation merging caused by conflicting BBox directions (step~\ref{alg:direction_s}-\ref{alg:direction_e}). 
At last, we merge BBoxes in each cluster to one single proposal by weighing the BBox parameters with their prediction confidence $s_i$ (step~\ref{alg:merge_s}-\ref{alg:merge_e}). 
After merging the BBoxes in each cluster, we end up with $K$ merged proposals, which are collected in the set $M$.

\begin{algorithm}[!h]
\caption{Cooperative BBox Matching}\label{alg:cbp}
\begin{algorithmic}[1]
\Ensure $\mathcal{B}=B_1\cup B_2\cup ...\cup B_{N_\text{v}}$, cluster set $\mathcal{C}=\emptyset$, cluster index $k=1$,  $\mathrm{iou_{thr}}=0.3$, merged proposal set $M=\emptyset$.

\While{$ \mathcal{B} \neq \emptyset $} \label{alg:clustering_s}%
    \State Select one BBox $b$ from set $\mathcal{B}$
    \State $\mathrm{C_k}=\{(b', s')|(b', s')\in \mathcal{B},  \text{IoU}(b', b)>\mathrm{iou_{thr}}\}$ 
    \State $\mathcal{B}\leftarrow \mathcal{B}\setminus\mathrm{C_k}$
    , $\mathcal{C}\leftarrow \mathcal{C}\bigcup \{\mathrm{C_k}\}$
    , $k = k+1$
\EndWhile    \label{alg:clustering_e}

\ForEach {$ \mathrm{C_k} \subset \mathcal{C}$}
    \State $I \leftarrow \{i \mid (s_i, b_i) \in \mathrm{C_k}\}$
    \State $r_{\text{max}}=$ ${\mathrm{argmax}_r}$\, $S$,\, $S=\{s_{i}|i\in I\}$ \label{alg:direction_s}
    
    \State $S_\text{dir1}=\sum\limits_{I_1} s_{i1}, \,{I_1}=\{i1 \mid \lvert r_{i1} - r_{\text{max}}\rvert_{a}>\frac{\pi}{2}, i1 \in I\}$
    \State $S_\text{dir2}=\sum\limits_{I_2} s_{i2}, \,{I_2}=\{i2 \mid \lvert r_{i2} - r_{\text{max}}\rvert_{a}\leq\frac{\pi}{2}, i2 \in I\}$\quad
    \Comment{$\lvert \cdot \rvert_{a}$ is the angle difference and normalized to $[0, \pi]$}
    \State $I_\text{max}\leftarrow \argmax_{\{I_1, I_2\}}\, (S_\text{dir1}, S_\text{dir2})$
    \ForAll {$i \in I_\text{max}$} \,$r_{i} \leftarrow r_{i} + \pi$\nobreak\EndFor\label{alg:direction_e} 
    \State $s_{i,\text{norm}}=s_i / \sum\limits_{j} s_i$ , $i, j \in I$ \label{alg:merge_s}
    \State $m_{*} = \sum\limits_i b_{i*} \cdot s_{i,\text{norm}}$ , $*\in \{x,y,z,w,l,h\}$
    , $i \in I$
    \State $m_{r}=\arctantwo(\sum\limits_{i} s_{i,\text{norm}}\cdot \sin{r_{i}} , \sum\limits_{i} s_{i,\text{norm}}\cdot \cos{r_{i}}),\, i \in I$\label{alg:merge_e}
    \State $M\leftarrow M \cup \{(m_{x}, m_{y},m_{z}, m_{l}, m_{w}, m_{h}, m_{r})\}$
\EndFor

\State\Return $M$
\label{ag:Matching}
\end{algorithmic}
\end{algorithm}
\setlength{\dbltextfloatsep}{0pt}

As shown in Fig.~\ref{fig:fusion_pipeline}c, the merged proposals $M$ (black box) are refined by aggregating the information around this proposal, namely, the neighboring keypoints (darker colored points) coming from different CPMs (blue and orange). 
This aggregation is achieved by a VSA-based RoI-grid pooling module which is originally proposed by \cite{pvrcnn}. 
It divides the proposal box into regular grids and summarizes the neighboring keypoints information for each grid center. 
The aggregated grid features are then stretched to a vector and fed to the fully connected layers to generate the final cooperative detection result which contains a binary classification between positive and negative proposals and the proposal box refinement regression. 
Different to~\cite{pvrcnn}, we replaced the batch normalization (BN)~\cite{batch_norm} in the fully connected layers with dropout~\cite{dropout}. 
Because of the computational overhead of multiple point clouds in each frame, we are only able to set the batch size to one during training, which does not satisfy the condition of BN.

\subsection{CPM compression}
We follow~\cite{v2vnet} to compress the encoded CPM features using Draco~\footnote{{DRACO}: 3D data compression. \url{https://google.github.io/draco/}}in order to take compression also into consideration when comparing the CPM size of sharing original feature maps and keypoint features. 
For both feature formats, we first write the 2D points of feature maps or the 3D keypoints to PLY~\footnote{{PLY}: Polygon File Format. \url{http://paulbourke.net/dataformats/ply}} file format and then compress this file with Draco.

\subsection{Localization error correction}
Since our 3D fusion model relies on highly accurate 3D keypoints, localization error will drastically reduce the performance of \textbf{FPV-RCNN}. 
To avoid this, a localization error correction module is introduced before the BBox matching (Algorithm \ref{alg:cbp}). 
Firstly, we add the semantic classification head upon the deep features of the selected keypoints as described in Fig.~\ref{fig:feature_selection}. 
Then, the keypoints are classified into classes of wall, fence, pole, vehicle, and others. 
Based on the semantic classes, we select all $K_\text{p}$ points of poles and $K_\text{fw}$ points of walls and fences through down-sampling with the FPS.
In addition to $C_i$, $B_i$ and $F_i$, only the x- and y-coordinate of the selected $K_i = K_\text{p} + K_\text{fw}$ points are shared to correct the localization error. 
This is described in the dialog box in Fig.~\ref{fig:fusion_pipeline} as the 3rd content of CPM. 
Based on the selected keypoints of poles, fences, walls and the vehicle centers, we use the maximum consensus algorithm~\cite{max_cons} with a rough searching resolution to find the corresponding vehicles centers and poles points, and then use these correspondences to calculate the accurate error estimation. 
We do not use wall and fence points for the final error calculation because matching on them leads to inaccurate result.

\section{Experiments}
\label{sec:Experiments}

\subsection{Dataset}
\label{dataset}
To evaluate the performance of the proposed method, we use a synthetic cooperative perception dataset called COMAP~\cite{comap}, which is simulated by CARLA \cite{carla} and SUMO \cite{sumo}. 
Many existing real-world datasets, \eg~KITTI~\cite{kitti}, nuScenes~\cite{nuscenes2019}, and Waymo~\cite{waymo}, are more suited for ego-perception, whereas collective perception requires multiple CAVs to observe the same scene simultaneously with enough FOV overlaps. 
On the contrary, the synthetic dataset containing various realistic cooperative vehicle scenarios with accurate ground truth information is easy to acquire and needs no further manual work of data labeling.
In addition, the lack of benchmark datasets leads to difficulty in comparing the performance of different fusion methodologies. 
Hence, in this paper, we follow many other works~\cite{marvasti2020feature,lidarsim,v2vnet} to use such a synthetic dataset for the empirical studies. 

In total, there are 7788 frames of samples in COMAP---4155 frames for training and 3633 frames for the test. Each frame contains the point cloud from an ego vehicle, the point clouds from the cooperative vehicles in the ego vehicle's communication range within \SI{40}{m}, and the corresponding GT BBoxes of each CAV. 
The GT BBoxes are selected according to the detection range \SI{57.6}{m} as the same in~\cite{comap} to guarantee a minimum safety distance for an emergency brake. 
For communication efficiency, only up to four point clouds of the cooperative vehicles are loaded. 
To facilitate the feature fusion step, the orientation of all the point clouds is aligned to the world coordinate system. 
Besides, the z-coordinates (heights) are also aligned to avoid a big performance drop of the object detection caused by the LiDARs mounted on vehicles of different heights. 
After this alignment, all the point clouds are filtered by the detection range on the x-y plane and the height range $[-0.1, 3.9]$\SI{}{m}. 
During training, the occluded GT BBoxes with no observed reflected points are removed. 
In the end, the pre-processed point clouds are voxelized to a size of \SI{0.1}{m} before they are fed to the DNNs in the framework (see Fig.~\ref{fig:fusion_pipeline}).

\subsection{Comparative model and baseline}
\begin{figure}[t]
\begin{center}
    \includegraphics[trim=0.0in 0.0in 0.0in 0.0in, clip=true, width=0.37\textwidth]{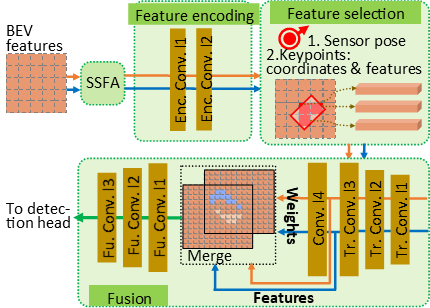}
    \caption{An overview of BEV deep feature fusion.}
\label{fig:bev_fusion}
\end{center}
\end{figure}

\paragraph{BEV keypoints deep features fusion}
Since the works that fuse deep features mentioned in Sec.~ \ref{related_works} all share BEV features, we also build a comparative model for the BEV feature fusion. 
However, different from previous works, we only select features that are inside the proposals $B_i$ for sharing to ensure a fair comparison between the BEV and 3D feature fusion with a similar magnitude of CPM size. 
We notate this framework as \textbf{BEV}.
The pipeline of BEV feature fusion is compatible with the one depicted in Fig.~\ref{fig:fusion_pipeline}a-c. The details of the modules that are different from \textbf{FPV-RCNN} are shown in Fig.~\ref{fig:bev_fusion}. 
The BEV features generated by feature extraction are passed to a Spatial-Semantic Feature Aggregation (SSFA) \cite{CIASSD} module, which can extract more robust features for generating accurate predictions. 
This feature map is further encoded and compressed by two convolutional layers and then selected by the proposals $B_i$. 
In addition to the selected BEV keypoints, the CPMs in this case also contain the sensor pose but no proposals because they are not needed for a single-stage detector.
In the fusion step, the shared feature maps are first up-sampled to a higher resolution by several transposed convolution layers and then merged by a summation of weighted feature maps. 
The weights are automatically adaptable as they are learned by a convolutional layer. 
The merged feature maps are then further fused and contracted by three convolutional layers to the detection resolution for the final detection. 

\paragraph{Baseline}
We take the raw data fusion strategy as a baseline.  
This strategy avoids any data loss during sharing, hence is more likely to perform best. 
Namely, two corresponding raw data fusion networks are taken as baselines---one for BEV-keypoints fusion (noted as \textbf{B\textsubscript{bev}}) and another for 3D-keypoints fusion (noted as \textbf{B\textsubscript{fpvrcnn}}). 
\textbf{B\textsubscript{bev}} takes CIA-SSD as the base object detector. 
Its fusion framework is adopted from~\cite{comap} and is partially taken from the \textbf{FPV-RCNN} framework that only contains the feature extraction and ego-detection module. 
For \textbf{B\textsubscript{fpvrcnn}}, we add VSA and RCNN (RoI-grid pooling and detection head) module to \textbf{B\textsubscript{bev}} to refine the proposals as similar to \textbf{FPV-RCNN} as possible.

\subsection{Experiment setup} \label{experiment_setup}
\paragraph{Training setting}
The targets for training are generated relative to the pre-defined anchors. 
For \textbf{B\textsubscript{bev}}, \textbf{BEV}, the ego detection of \textbf{B\textsubscript{fpvrcnn}}, and \textbf{FPV-RCNN}, we generate two anchors respective to rotations 0 and $\pi/2$ on each location of the $8\times$ down-sampled feature maps.
These anchors are of $[4.41, 1.98, 1.64]$\SI{}{m} in length, width, and height. 
An anchor is defined as positive if its IoU against the GT BBox is over 0.6, negative if under 0.45, and is ignored otherwise for the classification.
For the cooperative detection of \textbf{B\textsubscript{fpvrcnn}} and \textbf{FPV-RCNN}, we generate targets relative to the merged proposals by the ego detection (see Algorithm~\ref{alg:cbp}). 
But a single IoU threshold of 0.3 is used to separate the positive ($\geq0.3$) and negative ($<0.3$) samples. 
For the ego detection, we supervise the prediction results of all the incoming point clouds $\textbf{PC}$ (see Sec.~\ref{sec:formulation}) separately. However, for the cooperative perception, we only supervise the detection results from the perspective of the ego point cloud.

The same loss functions and parameters for SSD head from the original work~\cite{CIASSD} are adopted for object classification. 
But positive and negative samples are weighted differently, \ie~50 vs. 1, to prevent the network from classifying all samples as negative. 
For the RCNN head, a binary cross-entropy loss is used for classification and a smooth $\text{L}_1$-loss for regression. 
They are normalized over all samples.
Since the CAVs also share their own poses with each other, we also add the GT BBoxes of the ego vehicle and all the selected cooperative vehicles to the detection before feeding them to the Non-Maximum-Suppression (NMS). 
The thresholds for the classification scores and the NMS IoUs are set to 0.3 and 0.01, respectively, and kept the same in the test phase.

We run all the experiments only on a single Nvidia 1080Ti GPU to simulate a restricted computational resource in an AV.
\textbf{B\textsubscript{bev}} is trained from scratch for 50 epochs with a batch size of 8 frames. 
The trained weights are used for initializing the weights of the feature extraction and ego detection module in \textbf{B\textsubscript{fpvrcnn}}, \textbf{BEV}, \textbf{FPV-RCNN}.
These three networks are then further fine-tuned for 10 epochs with a batch size of 4 for \textbf{B\textsubscript{fpvrcnn}} and 1 for the other two.
The Adam optimizer (coefficients of $0.95 \& 0.999$) is applied to optimize the losses by stochastic gradient descent.
Its learning rate and decay both are set to $1e^{-4}$. 
We provide the detailed settings in our \href{https://github.com/YuanYunshuang/FPV\_RCNN}{\texttt{repository}} for reproducing our models.

\paragraph{Test setting}
Different numbers of cooperative vehicles are tested for analyzing the performance of cooperative perception.
This is done by fixing the number of cooperative vehicles $N_\text{v}$ in each test run. 
Namely, $N_\text{v}$ varies from 0, 2 to 4. In each run, only the frames having at least $N_\text{v}$ cooperative point clouds are selected as a test set for evaluation. 
If there are more than $N_\text{v}$ cooperative point clouds, we randomly select $N_\text{v}$ out of them to simulate the random geometric distribution of CAVs. 

Moreover, different CPM feature channels are analyzed for the keypoints feature fusion.
We set CPM feature channels to 128 to compare with both the BEV and 3D keypoints fusions under the condition of no information loss during the CPM compression process.
To further investigate the possibility of decreasing the size of CPMs in the \textbf{FPV-RCNN} framework, we conduct a series of experiments by setting different $N_\text{kpts}$ for FPS (2048 and 1024) and different CPM feature encoding channels $N_\text{ch}$ (128, 64, and 32).

Maximum consensus algorithm is very stable against the magnitude of the noise---different noise distributions only lead to a change in the search range of the maximum consensus algorithm. Therefore, we only use one fixed normal distribution for the absolute localization error of each vehicle to investigate the influence of pose errors on the fusion framework. This is different from \cite{v2vnet} which imports errors to the relative pose between ego and cooperative vehicles and vary the translation error from 0 to \SI{0.4}{m}, the rotation error from 0 to $4^\circ$. In our experiment we only use the biggest error setting from their work for the global localization error of both ego and cooperative vehicles: $\mathcal{N}(0, 0.4^2)m$ in x- and y-direction and $\mathcal{N}(0, 4^2)^{\circ}$ for the orientation of the vehicles. This will lead to much larger relative errors.
According to the error standard deviation, the search range of maximum consensus is empirically set to $[-1, 1]m$ for x- and y-axis and $[-6, 6]^{\circ}$ for the orientation. The searching resolution is set to \SI{1}{m} and $1^{\circ}$ for the translation and orientation, respectively.

\paragraph{Evaluation metrics}
All results are evaluated by Average Precision (AP) defined by the Area Under Precision-Recall Curve. 
IoU criteria $(0.3, 0.5, 0.7)$ are used for counting the positive detection to evaluate the detection performance. 

\section{Result and Evaluation}
\label{sec:ResultandEvaluation}
\subsection{Comparison with baselines}
\renewcommand{\arraystretch}{1}
\begin{table}[t!]
    \centering
        \caption{AP of different fusion models (in \%)}
    \begin{tabular}{c|c||c|c|c|c}
    \toprule
        IoU &  $N_\text{v}$ & \cellcolor{gray!10}\textbf{B\textsubscript{bev}} & \cellcolor{gray!10}\textbf{B\textsubscript{fpvrcnn}} & \textbf{BEV} & \textbf{FPV-RCNN} \\\midrule
        \multirow{3}{*}{0.3} 
        & 0 & \cellcolor{gray!10}\textbf{72.78} & \cellcolor{gray!10} 69.20 & \textbf{74.02}  & 67.83 \\
        & 2 & \cellcolor{gray!10}\textbf{90.45} & \cellcolor{gray!10} 88.77 & \textbf{89.73}  & 89.20 \\
        & 4 & \cellcolor{gray!10}\textbf{95.74} & \cellcolor{gray!10} 94.64 & 94.32  & \textbf{94.61}\\\hline
        \multirow{3}{*}{0.5}            
        & 0  & \cellcolor{gray!10}\textbf{70.42} & \cellcolor{gray!10} 67.76 & \textbf{71.62} & 66.08  \\
        & 2 & \cellcolor{gray!10}\textbf{89.11} & \cellcolor{gray!10} 87.84 & 88.32 & \textbf{88.39}\\
        & 4 & \cellcolor{gray!10}\textbf{94.78} & \cellcolor{gray!10} 94.06 & 93.07 & \textbf{94.02}\\\hline
        \multirow{3}{*}{0.7}
        & 0 & \cellcolor{gray!10}57.98 & \cellcolor{gray!10} \textbf{58.73} & \textbf{61.59} & 58.37\\
        & 2 & \cellcolor{gray!10}\textbf{82.94} & \cellcolor{gray!10} 82.75 & 74.60  & \textbf{83.79}\\
       &  4 & \cellcolor{gray!10}\textbf{90.73} & \cellcolor{gray!10} 90.70 & 82.21  & \textbf{90.88}\\ \bottomrule
    \end{tabular}
    \label{tab:result1}
\end{table}
\renewcommand{\arraystretch}{1}

Table~\ref{tab:result1} shows the AP scores of the baselines (in the gray cell) and the fusion models. 
With cooperative vehicles ($N_\text{v}>0$), compared to the \textbf{B\textsubscript{bev}} and \textbf{B\textsubscript{fpvrcnn}} fusion baselines (bold font in the gray cell), BEV-fusion has an acceptable small performance drop at the low IoU threshold (0.3). It is worth noting that the performance of \textbf{FPV-RCNN} even surpasses that of \textbf{B\textsubscript{fpvrcnn}} with a small AP gain at different IoUs. 
For example, when there are only two cooperative vehicles, the AP of \textbf{BEV} at $\text{IoU}=0.3$ drops 0.72\% while that of 3D-fusion even increases 0.43\%, compared to their respective baselines. 
However, as the IoU threshold increases to 0.5, the gap between \textbf{BEV} and \textbf{B\textsubscript{bev}} slightly increases. 
At $\text{IoU}=0.7$, the gap between them even increases to 8.34\%. 
In contrast, the performance of \textbf{FPV-RCNN} is slightly better than its baseline \textbf{B\textsubscript{fpvrcnn}}, and their performance gaps are small and remain consistent. 
This implies that the additional RCNN-head helps improve the localization accuracy of the BBoxes at lower IoU thresholds, but not the recall of the BBoxes.  
This is because the RoI-grid pooling can better aggregate the 3D keypoints features learned from the point cloud for high accurate BBox predictions. 
In other words, compared to \textbf{BEV}, our model is more suitable for feature fusion of cooperative object detection with respect to highly accurate and reliable BBox predictions.

Nevertheless, when there are no cooperative vehicles, our \textbf{FPV-RCNN} performs much worse than the other two baselines. 
This is because these self-dependent detection results are generated only by the feature extraction and ego-detection module. 
The weights of these two modules are fine-tuned during the training of the whole BEV and 3D fusion framework. 
This observation indicates that \textbf{BEV} tends to learn features that are more helpful for detection tasks on a single point cloud. 
Thus, it overfits under such a configuration with better performance than that of the more generalized \textbf{B\textsubscript{bev}} (\eg~61.59\% vs. 57.98\% at $\text{IoU}=0.7$). 
In contrast, \textbf{FPV-RCNN} focuses on learning features that are useful for the later fusion, and therefore are counter-affected by the original pre-trained weights for non-cooperative detection. 
It should be noted that this issue can be circumvented by loading different pre-trained weights according to the requirements in real applications.

\subsection{FPV-RCNN performance with variate CPM sizes}\label{sec:result2}
Table~\ref{tab:result2} shows the results of \textbf{FPV-RCNN} with different CPM encoding parameters.
$N_\text{kpts}$ stands for the number of keypoints for FPS and $N_\text{ch}$ stands for the number of channels for encoding the CPM features. 
Besides, the results are evaluated with two different numbers of cooperative vehicles ($N_\text{v}=\{2, 4\}$). 
In general, the better performance is mostly associated with a larger $N_\text{kpts}$ and all best AP scores (bold blue font) appear when $N_\text{v}=2048$. 
But for a specific IoU and $N_\text{v}$,  the performance only varies within a range of less than 1\% for different $N_\text{ch}$.
Interestingly, in most cases, the best AP even appears at the smallest $N_\text{ch}$ (bold font).

\renewcommand{\arraystretch}{1}
\begin{table}[t!]
\setlength{\tabcolsep}{5pt}
    \centering
    \caption{Performance of FPV-RCNN with different number of keypoints and CPM encoding channels (AP in \%)}
    \begin{tabular}{c|c||c|c||c|c}
    \toprule
        \multicolumn{2}{c||}{ $N_\text{v}\rightarrow$} & \multicolumn{2}{|c||}{2}  & \multicolumn{2}{|c}{4}\\ \hline
         IoU$\downarrow$ & $N_\text{ch}\downarrow$, $N_\text{kpts}\rightarrow$  & 2048 & 1024 & 2048 & 1024 \\ \midrule
        \multirow{3}{*}{0.3} 
        & 128 & 89.20 & 88.74 & 94.61 & 93.89\\
        & 64  & 88.47 & 89.03 & 94.40 & 94.41\\
        & 32  & \textbf{\textcolor{blue}{89.93}} & \textbf{89.10} & \textbf{\textcolor{blue}{95.23}} & \textbf{94.59}\\\hline
        \multirow{3}{*}{0.5}
        & 128 & 88.39 & 87.80 & 94.02 & 93.15\\
        & 64  & 87.79 & 88.11 & 93.85 & 93.76\\
        & 32  & \textbf{\textcolor{blue}{88.72}} & \textbf{88.15}  & \textbf{\textcolor{blue}{94.23}}  & \textbf{93.91}\\\hline
        \multirow{3}{*}{0.7}
        & 128 & \textbf{\textcolor{blue}{83.79}} & \textbf{83.29} & {90.88} & 89.96\\
        & 64  & 83.10 & 83.14 & 90.62 & 90.12 \\
        & 32  & 83.76 & 83.25 & \textbf{\textcolor{blue}{90.99}} & \textbf{90.62}\\ \bottomrule
    \end{tabular}
    \label{tab:result2}
\end{table}
\renewcommand{\arraystretch}{1}

\begin{figure}[ht!]
\begin{center}
    \includegraphics[trim=0.0in 0.0in 0.0in 0.2in, clip=true,width=0.45\textwidth]{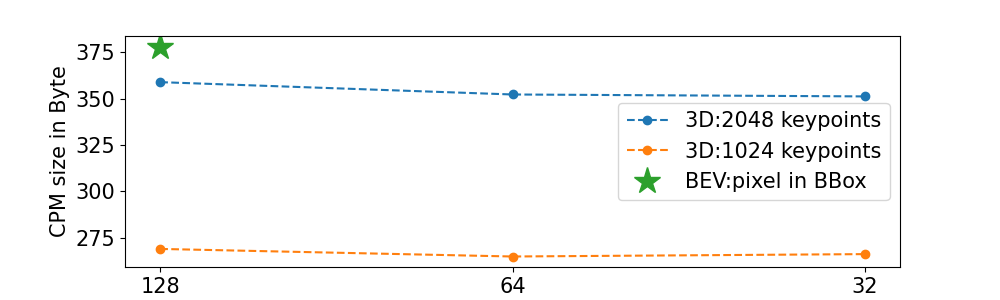}
    \caption{CPM size comparison}
\label{fig:cpm_sizes}
\end{center}
\end{figure}

Furthermore, we compare the CPM sizes of the compressed deep features averaged over all CPMs and CAV numbers. 
Fig.~\ref{fig:cpm_sizes} gives a quantitative comparison between BEV and 3D (FPV-RCNN, noted with a different $N_\text{kpts}$) keypoints feature sharing.
It can be seen that the average CPM size of the compressed keypoints features is decreased to around 0.3KB, which is about 50 times smaller than the CPM generated by compressing the whole feature maps (ca. 14KB).
With the same number of feature channels ($N_\text{ch}=128$), 3D keypoints fusion transmits less data than the BEV keypoints fusion but achieves an enhanced performance by a big margin ($90.88\%$ vs. $82.21\%$, see Table~\ref{tab:result1}).
Besides, our framework also generates CPMs with sizes in the same order of magnitude as the object-based standardized CPM~\cite{ETSI} evaluated in a low traffic density scenario by~\cite{ CP_florian}.

These observations above indicate that our proposed framework is relatively stable against the variation of the CPM feature encoding size. 
Hence, if the communication network is not fully consumed and the wireless network can handle larger CPMs, it is preferable to increase $N_\text{kpts}$ rather than increase the feature encoding channels. 
On the other hand, as a big advantage, if the communication network is already heavily loaded, the CPMs can be compressed as small as possible with only a slight performance drop. 

\subsection{Ablation study with respect to localization noise}
\begin{figure}[ht!]
\begin{center}
    \includegraphics[trim=0.35in 0.0in 0.8in 0.6in, clip=true,width=0.35\textwidth]{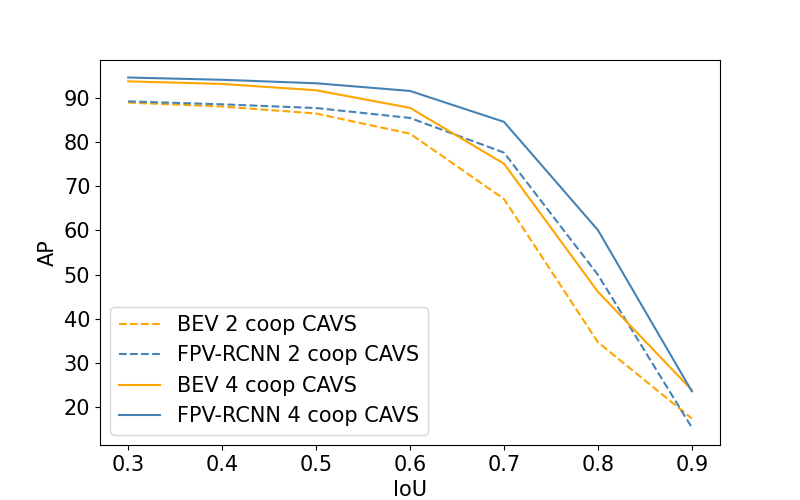}
    \caption{Results with localization errors.}
\label{fig:loc_noise1}
\end{center}
\end{figure}

In order to show the effectiveness of the matching module we proposed in Algorithm~\ref{alg:cbp}, we compare the result of this module with the NMS object fusion used in V2Vnet~\cite{v2vnet}. 
As shown in the bold font in Table~\ref{tab:ablation}, our matching module outperforms the NMS fusion in most of the cases. 
Especially, when localization noise exists, matching using Algorithm~\ref{alg:cbp} is more stable. 
In addition, we also studied the performance gain of the 3D feature fusion of \textbf{FPV-RCNN} in the second stage. 
The best-performed results are marked in blue, which clearly indicates that our fusion module of \textbf{FPV-RCNN} can refine the results. 
Moreover, by observing the detection results, we found that most of the false positive detection is removed by RCNN in the second 3D feature fusion stage. 
However, this effect can hardly influence the result of AP. 
Therefore, we do not observe large AP improvement between the results of the full \textbf{FPV-RCNN} and Algorithm~\ref{alg:cbp}.
Moreover, as shown in Table~\ref{tab:ablation}, our \textbf{3D} model with localization noise, as expected, performs worse than the no-noise version. But the performance at lower IoU thresholds is only slightly dropped. However, as shown in Fig.~\ref{fig:loc_noise1}, with localization noise, our \textbf{3D} model still performs better than the BEV fusion model. Since the raw data fusion baselines do not have the keypoints selection and classification module, it is difficult to correct localization error and their results are not plotted in Fig.~\ref{fig:loc_noise1}.
\renewcommand{\arraystretch}{1}
\begin{table}[ht!]
\setlength{\tabcolsep}{3.5pt}
    \centering
    \caption{Ablation study with and without noise (AP in \%)}
    \begin{tabular}{c|c||c|c|c||c|c|c}
    \toprule
         \multicolumn{2}{c||}{-}  & \multicolumn{3}{|c||}{no noise}  & \multicolumn{3}{|c}{with noise}\\ \hline
         $N_\text{v}$ & IoU & NMS & Alg.~\ref{alg:cbp} & \textbf{FPV-RCNN} & NMS & Alg.~\ref{alg:cbp} & \textbf{FPV-RCNN} \\ \midrule
         \multirow{3}{*}{2} 
         &0.3 & 88.08 & \textbf{88.73} & \textbf{\textcolor{blue}{89.20}} & 87.88 & \textbf{88.59} & \textbf{\textcolor{blue}{89.26}}  \\
         &0.5 & 86.89 & \textbf{87.32} & \textbf{\textcolor{blue}{88.39}} & 86.26 & \textbf{86.58} & \textbf{\textcolor{blue}{87.32}} \\
         &0.7 & \textbf{82.57} & 81.94 & \textbf{\textcolor{blue}{83.79}} & 77.06 & \textbf{\textcolor{blue}{78.13}} & 77.86 \\
         \hline
         \multirow{3}{*}{4}
         &0.3 & 93.47 & \textbf{94.01} & \textbf{\textcolor{blue}{94.61}} & 93.34 & \textbf{93.75} & \textbf{\textcolor{blue}{94.70}} \\
         &0.5 & 92.54 & \textbf{92.85} & \textbf{\textcolor{blue}{94.04}} & 91.97 & \textbf{92.28} & \textbf{\textcolor{blue}{93.21}} \\
         &0.7 & \textbf{89.51} & 88.96 & \textbf{\textcolor{blue}{90.88}} & 83.75 & \textbf{85.31} & \textbf{\textcolor{blue}{85.54}}\\
    \bottomrule     
    \end{tabular}
    \label{tab:ablation}
\end{table}
\renewcommand{\arraystretch}{1}


Nevertheless, the current model also has several limitations. First, we only carried out the empirical studies on the synthetic data. 
In our future work, first we will extend our experiment to real-world data to further analyze the efficacy of the proposed FPV-RCNN model. 
Second, the communication delay is only reflected by the CPM sizes. This needs to be further investigated by analyzing the CPM communication and transmission in real-world scenarios.

\section{Conclusion}
In this paper, we proposed an efficient framework, called \textbf{FPV-RCNN}, for point cloud-based cooperative vehicle detection of autonomous driving. 
The framework takes PV-RCNN~\cite{pvrcnn} as the base network of object detection for cooperative perception scenarios by adding a keypoints selection module, a bounding box proposal matching module with localization error correction, and the keypoints fusion module.
The comparison to a 2D BEV feature fusion on a simulated dataset COMAP~\cite{comap} shows that our method improves the performance of cooperative vehicle detection by a big margin. 
In comparison to previous works that share full BEV feature maps, our method significantly decreases the data transmission load in the CAV network for real-time communication and is also more robust against localization noise thanks to the noise correction module. 
In future work, we plan to evaluate our method in real-world cooperative driving scenarios.

\section*{Acknowledgment}
This work is supported by the projects \href{www.socialcars.org}{DFG RTC1931 SocialCars} and \href{https://www.icsens.uni-hannover.de/}{DFG GRK2159 i.c.sens}.

\bibliography{ref, IEEEabrv}

\end{document}